\documentclass[times,review,10pt]{elsarticle}

\usepackage{amssymb}

\usepackage[latin9]{inputenc}
\usepackage[T1]{fontenc}
\usepackage[english]{babel}
\usepackage{graphicx}
\usepackage{rotating}
\usepackage{array}
\usepackage{float}
\usepackage{subfigure}
\usepackage{caption}
\usepackage{xcolor}
\usepackage{mathrsfs}
\usepackage{setspace}
\usepackage{epstopdf}
\usepackage{comment}
\usepackage{amsmath}
\usepackage{amssymb}
\usepackage{graphics} 
\usepackage{booktabs}
\usepackage{colortbl}
\usepackage{arydshln}
\usepackage{multicol}
\usepackage{xspace}
\usepackage{makecell}
\usepackage{tabularx}
\usepackage{bm}
\usepackage{pgfplots}

\usepackage{algpseudocode}

\usepackage[ruled,linesnumbered]{algorithm2e}
\usepackage{bbding} 
\usepackage{multirow}

\usepackage[normalem]{ulem}
\usepackage[safe]{tipa}

\usepackage{xcolor}

\SetCommentSty{mycommfont}
\usepackage{mathtools}

\usepackage{graphicx}
\usepackage{subcaption}

\usepackage{wrapfig}
\usepackage{flushend}

\journal{Pattern Recognition}

\pgfplotsset{compat=1.18}
\begin{document}

\begin{frontmatter}



\title{EmoDiffusion: Enhancing Emotional 3D Facial Animation with Latent Diffusion Models}

\author{Yixuan Zhang\fnref{label1}}
\author{Qing Chang\fnref{label2}}
\author{Yuxi Wang\fnref{label1}}
\author{Guang Chen\fnref{label3}}
\author{Zhaoxiang Zhang\fnref{label1,label2,label4}}
\author{Junran Peng\fnref{label3}}

\affiliation[label1]{organization={CAIR},,
            city={Hong Kong},
            postcode={999077}, 
            state={Hong Kong},
            country={China}}

\affiliation[label2]{organization={UCAS},
            city={Beijing},
            postcode={100190}, 
            state={Beijing},
            country={China}}

\affiliation[label3]{organization={USTB},
            city={Beijing},
            postcode={100083}, 
            state={Beijing},
            country={China}}

\affiliation[label4]{organization={The New Laboratory of Pattern Recognition (NLPR)},
            city={Beijing},
            postcode={100190}, 
            state={Beijing},
            country={China}}



\begin{abstract}
   Speech-driven 3D facial animation seeks to produce lifelike facial expressions that are synchronized with the speech content and its emotional nuances, finding applications in various multimedia fields. However, previous methods often overlook emotional facial expressions or fail to disentangle them effectively from the speech content. To address these challenges, we present EmoDiffusion, a novel approach that disentangles different emotions in speech to generate rich 3D emotional facial expressions. Specifically, our method employs two Variational Autoencoders (VAEs) to separately generate the upper face region and mouth region, thereby learning a more refined representation of the facial sequence. Unlike traditional methods that use diffusion models to connect facial expression sequences with audio inputs, we perform the diffusion process in the latent space. Furthermore, we introduce an Emotion Adapter to evaluate upper face movements accurately. Given the paucity of 3D emotional talking face data in the animation industry, we capture facial expressions under the guidance of animation experts using LiveLinkFace on an iPhone. This effort results in the creation of an innovative 3D blendshape emotional talking face dataset (3D-BEF) used to train our network. Extensive experiments and perceptual evaluations validate the effectiveness of our approach, confirming its superiority in generating realistic and emotionally rich facial animations.

\end{abstract}






\begin{keyword}
3D talking face \sep Diffusion model \sep Emotional 3D talking face

\end{keyword}

\end{frontmatter}


\section{Introduction}
Virtual characters and intelligent avatars have undergone significant advancements in recent years, becoming increasingly integral to multimedia applications across various industries, including commerce~\cite{wohlgenannt2020virtual}, entertainment~\cite{ping2013computer}, and education~\cite{tanaka2022acceptability}. These advancements have driven demand for more realistic and expressive 3D facial animations, a critical component for creating believable characters. However, 3D facial animation remains a resource-intensive process, requiring meticulous frame-by-frame adjustments to achieve high-quality results. Such manual efforts are both time-consuming and costly, with the adjustment of a single keyframe often taking hours of detailed work. This limitation poses a significant bottleneck for the widespread adoption of high-fidelity facial animation in applications that require scalability and efficiency.

To address these challenges, 3D speech-driven facial animation technologies have emerged as a promising solution. By leveraging voice assistants and interaction systems, these technologies enable the automated creation of natural and dynamic 3D facial animations tailored for remote communication and real-time interaction~\cite{pham2017speech}. The rapid evolution of deep learning has further empowered both industry professionals and researchers to embrace end-to-end speech-driven facial generation techniques~\cite{karras2017audio}. These methods democratize the production of high-quality 3D facial animations, significantly reducing both the time and cost associated with traditional animation workflows. Consequently, these technologies are transforming multimedia applications, bridging the gap between creative demands and practical constraints.

\begin{figure}[!tbp]
\centering
\includegraphics[width=0.5\textwidth, trim=100 0 80 0,clip]{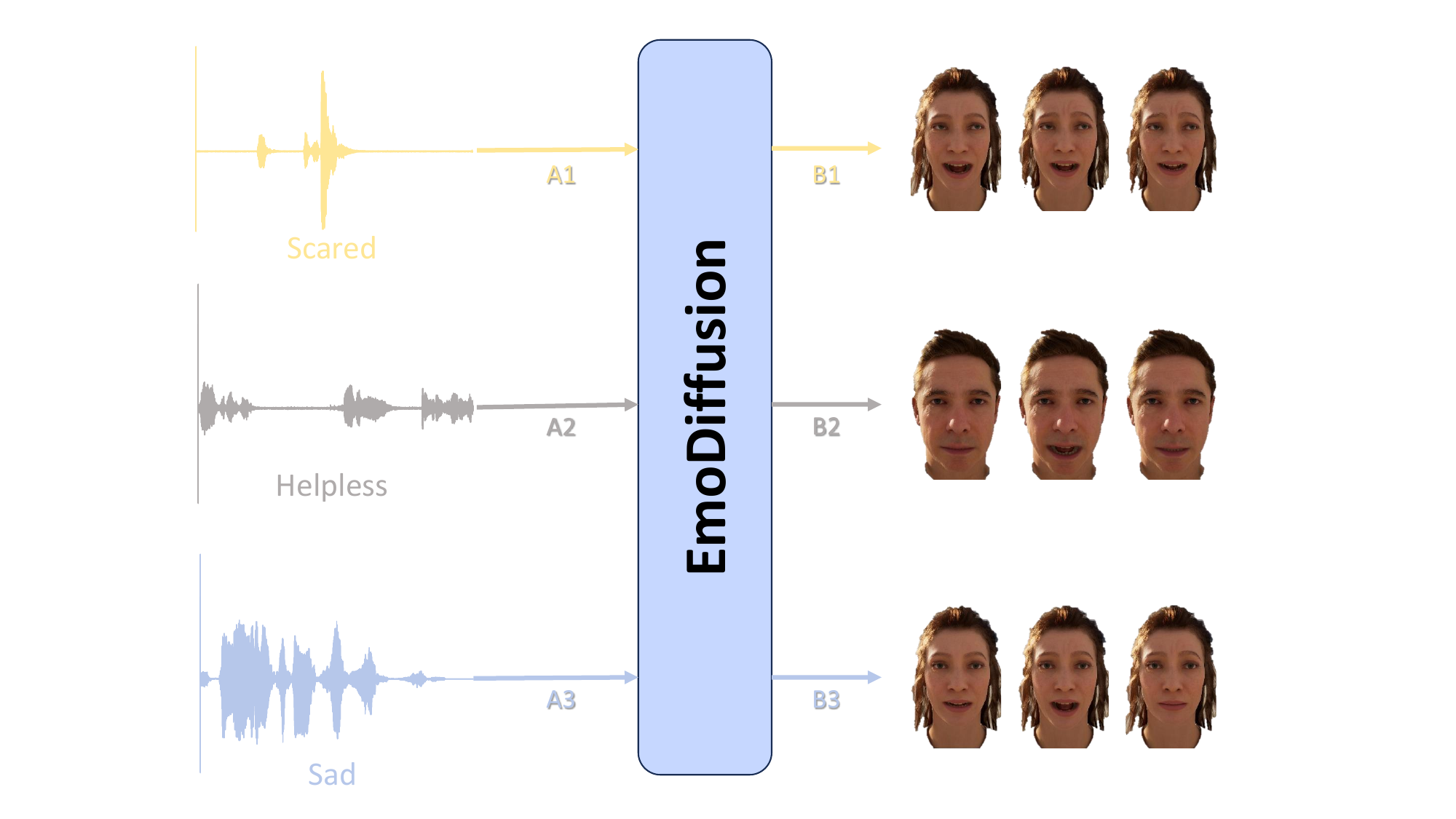}
\caption{Given audio input featuring diverse emotional tones, EmoDiffusion generates natural and vivid 3D facial expression sequences that accurately reflect the corresponding emotions.}
\label{fig1}
\end{figure}

Recent advancements in speech-driven 3D facial animation can be categorized into two primary approaches: \emph{lip-synchrony facial animation} and \emph{emotion-enhanced facial animation}. Lip-synchrony facial animation focuses on achieving precise alignment between spoken words and mouth movements, ensuring natural speech articulation. For example, Codetalker~\cite{xing2023codetalker} utilizes a Vector Quantized Variational AutoEncoder (VQ-VAE)~\cite{van2017neural} to effectively model facial actions, while SelfTalk~\cite{peng2023selftalk} employs a cross-modal network, refined by lip supervision, to produce coherent 3D talking faces.

In contrast, emotion-enhanced facial animation integrates emotional cues from audio inputs to enrich facial expressions and enhance character expressiveness. A notable example is Emotalk~\cite{peng2023emotalk}, which incorporates global emotional features into the animation process. However, Emotalk primarily focuses on high-level emotional attributes, often neglecting the nuanced facial movements---such as subtle eyebrow raises or eyelash flickers---that are essential for achieving vivid and lifelike talking head animations.

The advent of generative models, particularly diffusion-based approaches, has introduced a transformative paradigm in speech-driven facial animation. Diffusion models excel at generating diverse and non-deterministic text-to-body movements~\cite{chen2023executing}, enabling more expressive and realistic facial animations. These models capture the intricate subtleties of human expressions and emotional transitions, elevating the overall animation quality. However, their effectiveness hinges on two critical factors: the \emph{scale and quality of training data} and the \emph{diversity and accuracy of emotional expressions}.

Existing public datasets, such as VOCASET~\cite{cudeiro2019capture} and BIWI~\cite{fanelli20103}, are constrained in size and predominantly focus on audio-driven mouth shapes, with limited consideration for corresponding emotional expressions. To address these limitations, some methods~\cite{danvevcek2023emotional, peng2023emotalk} have leveraged 3D data extracted from 2D videos using motion capture techniques~\cite{xu2021high, liao2024hardmo}. However, video-based facial tracking frequently suffers from inaccuracies and a lack of consistent front-facing views, resulting in suboptimal data quality.

The creation of a comprehensive dataset featuring synchronized emotional speeches and corresponding 3D facial expressions is essential for advancing emotional 3D facial animation. However, to the best of our knowledge, no publicly accessible dataset fulfills these requirements. This lack of resources highlights a critical gap between realistic talking faces and animation-style emotional expressions, making this task a persistent challenge.

To overcome this limitation, we have developed the 3D Blendshape Emotional Talking Face Dataset (3D-BEF), a pioneering dataset containing over 2{,}000 sequences with synchronized audio and facial expression dynamics. The dataset, collected using the LiveLinkFace application integrated into an iPhone device under the guidance of professional animators, captures nine distinct emotional categories: \emph{neutral, angry, doubtful, surprised, happy, sad, scared, serious,} and \emph{proud}. By utilizing blendshape coefficients, the 3D-BEF dataset improves adaptability across diverse facial animations and introduces emotional expressiveness in the upper facial regions---an area previously overlooked in existing resources.

In this paper, we introduce EmoDiffusion, a novel framework designed to generate realistic and emotionally expressive 3D facial animations synchronized with audio inputs (see Fig.~\ref{fig1}). EmoDiffusion is tailored for the 3D-BEF dataset and excels in producing precise mouth shapes while significantly enhancing anime-style emotional reactions driven by audio cues. Our approach divides the face into two distinct regions---the upper face and the mouth area---and employs a dual facial movement Variational Autoencoder (VAE) architecture to decouple the generation of these regions.

To ensure emotional congruence with audio inputs, we integrate an \emph{audio-conditioned diffusion model} into the VAE's latent structure, enabling the generation of realistic and emotionally consistent facial features. Furthermore, we introduce an Emotion Adapter within the upper latent space to guide the generation process, ensuring the accuracy and fidelity of upper face movements (see Fig.~\ref{fig2}). Through rigorous quantitative experiments and qualitative assessments conducted by professional animators and animation enthusiasts, we demonstrate the efficacy of EmoDiffusion in generating lifelike emotional expressions.

The main contributions of our work are as follows:

\begin{enumerate}
    \item We propose EmoDiffusion, a novel training framework that leverages the latent space of diffusion models to generate speech-driven, emotionally expressive 3D facial animations. By incorporating an Emotion Adapter, our approach significantly improves the clarity, coherence, and fidelity of generated animations.

    \item We present the 3D Blendshape Emotional Talking Face Dataset (3D-BEF), a first-of-its-kind dataset featuring nine distinct emotional audio categories in animation style. This dataset addresses critical gaps in existing resources by introducing blendshape coefficients and emotional expressiveness in the upper face.

    \item We demonstrate that EmoDiffusion outperforms state-of-the-art methods in generating diverse and high-quality emotional expressions from audio inputs, showcasing its potential for real-world applications.
\end{enumerate}
\section{Related Works}

\subsection{Speech-driven 3D Facial Animation}

The domain of speech-driven animation has evolved significantly, offering groundbreaking techniques that span the creation of both 2D and 3D animated characters. Thanks to the widespread availability of 2D audio-visual datasets~\cite{zhang2021flow,}, the creation of 2D talking faces has achieved commendable results~\cite{bregler2023video}. However, these techniques are unsuitable for 3D character models, extensively utilized in 3D gaming and virtual reality interactions. So in this work, we focus on animating vivid 3D facial expressions. 

One of the most significant challenges in speech-driven animation is the limited availability of comprehensive datasets that capture a broad spectrum of emotions. For example, VOCA~\cite{cudeiro2019capture} leverages temporal convolutions and control parameters to animate characters from any speech input against a predefined character mesh. Its effectiveness, however, is largely limited to producing precise mouth movements. This limitation stems from the VOCASET~\cite{cudeiro2019capture} dataset's inadequate representation of upper facial movements. Similarly, the BIWI dataset~\cite{fanelli20103} offers an audio-visual corpus encompassing speech and facial expressions in the form of dense dynamic 3-D face geometries. Yet, it falls short by excluding critical facial components such as eyelids, eyebrows, and the inner mouth, restricting the depth of emotional expressions it can model.

To enhance the expression in facial animation, FaceFormer~\cite{fan2022faceformer} introduces a Transformer-based architecture that meticulously extracts audio cues relevant to the context, facilitating the generation of continuous facial animations in an autoregressive manner. This method not only broadens the applicability across different sources but also allows for subtler adjustments in mouth movements, offering an improvement over VOCA's capabilities. Conversely, EmoVOCA~\cite{nocentini2024emovoca}, aimed at training emotional expressions via audio and corresponding labels, splits the face into upper and lower halves to assess deviations from a neutral expression during emotion conveyance. Despite this, its enhancement of facial expressions remains limited, a consequence of its dependency on the VOCASET dataset, highlighting the ongoing need for richer datasets in advancing the realism of facial animations.

Several studies have been pivotal in advancing the domain of lip-syncing during speech, with notable examples including Codetalker~\cite{xing2023codetalker} and Selftalk~\cite{peng2023selftalk}, which have showcased impressive lip-syncing capabilities. However, the absence of upper facial movements in these models has led to animations that lack lifelikeness. Meshtalk~\cite{richard2021meshtalk} addresses this issue by introducing a method that distinguishes between audio correlated and audio uncorrelated movements through a cross-modality loss, thus facilitating the synthesis of audio uncorrelated movements such as blinking and eyebrow movements. This significantly enhances the realism of facial animations. In a similar vein, Sadtalker~\cite{zhang2023sadtalker} leverages 3D motion coefficients derived from 3D Morphable Models (3DMM) for facial animation generation, incorporating Expnet for learning facial emotion expressions and a Variational Autoencoder (VAE) for head movement generation. Furthermore, Emotalk~\cite{peng2023emotalk} introduces an end-to-end neural network specifically designed for the extraction of emotional information from audio. This model employs an emotion disentangling encoder to explore the depth of emotional characteristics within speech, enabling the vivid portrayal of emotional animations via vocal expressions. Despite these advancements in their emotion expression animation, they primarily concentrates on global emotional features, thereby neglecting the detailed facial movements that accompany specific emotions and the dynamic expressions evident during transitions between emotions.

\subsection{Generative Models in Facial animation}

Generative models have significantly advanced the field of facial generation tasks, covering both 2D~\cite{liu2020synthesizing} and 3D talking heads~\cite{zhao2024media2face}. These advancements have enabled the creation of increasingly lifelike facial animations, contributing to a wide range of applications from virtual avatars to human-computer interaction interfaces.

Initially, generative models for facial animation, such as Variational Autoencoders (VAEs)~\cite{karras2019style}, Generative Adversarial Networks (GANs)~\cite{yin2022styleheat}, and Neural Radiance Fields (NeRFs)~\cite{guo2021ad}, laid the groundwork for significant breakthroughs in producing realistic facial expressions. These models enabled the generation of high-quality, photorealistic 2D and 3D images with increasing control over expression and motion. However, the arrival of Diffusion Generative Models~\cite{sohl2015deep} marked a revolutionary shift, offering unprecedented capabilities in image synthesis. With innovations like Stable Diffusion~\cite{rombach2022high}, DALL-E 2~\cite{ramesh2022hierarchical} and clip~\cite{radford2021learning}, diffusion models have demonstrated superior versatility and quality, pushing the boundaries of what generative models can achieve in facial animation. Inspired by these advancements, recent approaches~\cite{thambiraja20233diface} have adapted diffusion models to facial animation tasks, enabling the generation of dynamic and highly expressive facial movements. Diffusion models provide a new paradigm in which subtle emotional nuances and complex temporal sequences of facial expressions can be modeled with high fidelity.

A particularly transformative approach is embodied by 3DiFACE~\cite{thambiraja20233diface}, which introduced a one-to-many framework for generating diverse facial animations from a single speech input. This model's ability to inject randomness and support motion editing has been a key innovation, especially for 3D facial animation tasks, even when trained on relatively small datasets. This flexibility has broad implications for creative industries, allowing for personalized and variable animation outcomes from limited data. Similarly,FaceTalk~\cite{aneja2023facetalk} excels in synthesizing high-fidelity 3D head motion sequences from audio inputs. It captures expressive details, including hair, ears, and subtle eye movements, by integrating speech signals with the latent space of a neural parametric head model, thus achieving temporally coherent motion sequences of exceptional quality.

Despite the remarkable successes of these diffusion-based models, significant challenges remain. One key issue is the susceptibility of these models to noise and temporal inconsistencies in raw facial expression data, often resulting in unnatural artifacts or jitter in the generated animations~\cite{chen2020comprises}. Furthermore, directly applying diffusion models to raw facial data~\cite{aneja2023facetalk, tevet2023human} poses substantial computational challenges. The high computational cost, coupled with slow inference speeds, limits the scalability and real-time application of these models. Addressing these challenges, we propose a novel approach: a facial expression latent-based diffusion model. By operating in the latent space rather than directly on raw facial data, this method reduces computational overhead while maintaining or even enhancing the generative quality of facial animations. This approach leverages recent advances in latent space modeling~\cite{chen2023executing, huang2024stablemofusion} to more efficiently capture the complex temporal and emotional dynamics of facial expressions. By focusing on latent representations, we aim to mitigate the computational demands traditionally associated with diffusion models, while also improving the fidelity of the generated animations, particularly in terms of emotional expressiveness and temporal coherence.

In conclusion, our latent-based diffusion approach represents a significant leap forward in the efficient and realistic rendering of facial expressions and movements. As generative models continue to evolve, this work contributes to the broader goal of creating lifelike, emotionally resonant avatars capable of interacting in real-time with users in diverse applications, ranging from entertainment to telepresence and beyond.

\section{Method}
\label{sec:method}

\begin{figure*}
\centering
\includegraphics[width=\textwidth]{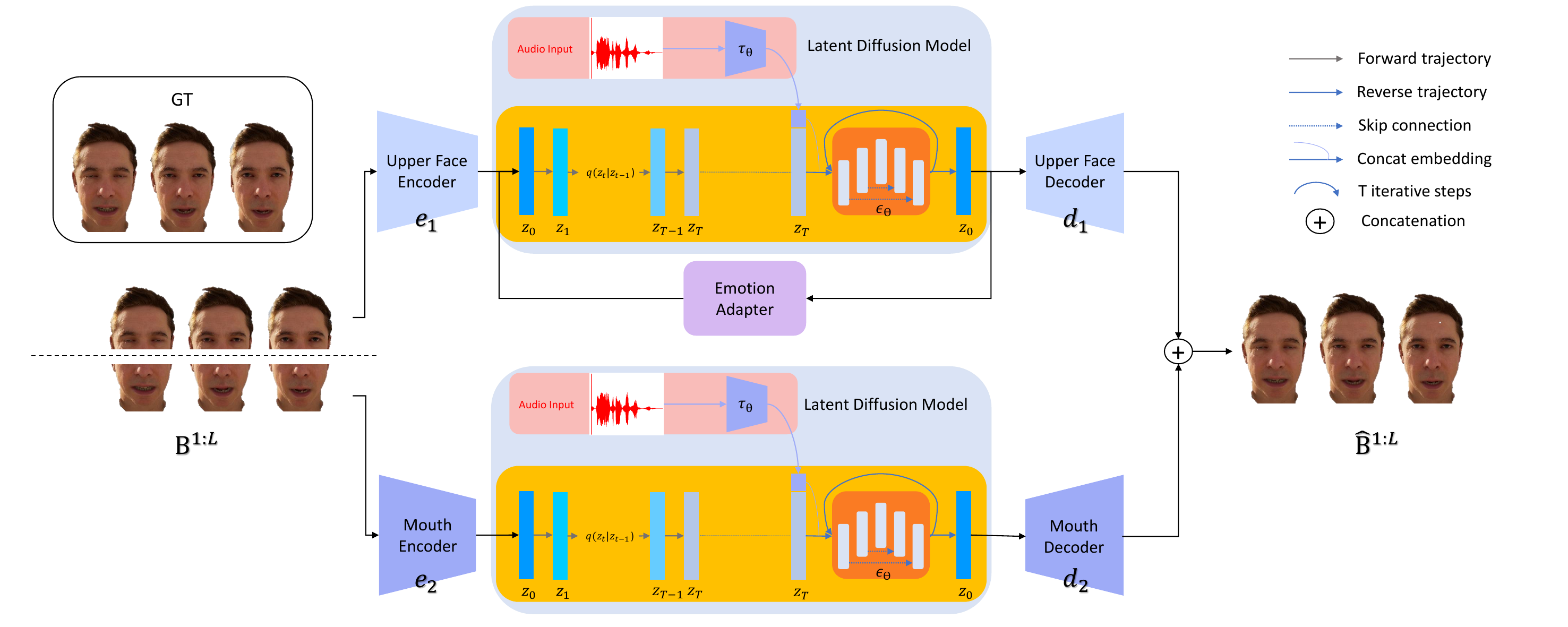}
\caption{Overview of EmoDiffusion. We propose EmoDiffusion, a novel framework that integrates two Variational Autoencoder (VAE) architectures using the facial blendshape coefficients data with their corresponding Latent Diffusion Models. This framework incorporates an Emotion Adapter-enhanced diffusion model specifically designed to augment the expressiveness of the upper face. Given an input audio signal, the audio encoder extracts an audio embedding, denoted as $\tau_{\theta}$. Utilizing the conditioned denoiser $\epsilon_{\theta}$, we reverse the diffusion steps to generate the latent facial expressions $\hat{z}_0$. These expressions are then decoded by the respective VAE decoders $\mathcal{D}$, and the outputs from each are concatenated to form the final facial expression sequence. This approach ensures a robust and expressive generation of facial movements synchronized with the input audio.}
\label{fig2}
\end{figure*}

To efficiently generate high-quality and vivid emotional facial expression sequences according to desired conditional inputs with minimal computational overhead, we propose EmoDiffusion, a dual Variational Autoencoder (VAE) structure latent diffusion model with an Emotion Adapter operating within the latent space. This synergistic architecture efficiently encodes face blenshapes sequences ($B^{1:L}$) into a low-dimensional yet richly diverse latent space ($Z$) , as delineated in Section~\ref{Dual Facial Expression Representation in Latents} of our work. By separating the face ($F$) into two regions: the upper part which contains eye regions ($f_1$) and the mouth area ($f_2$). We can constrain the upper face region to get the desired emotional result corresponding to the audio. Furthermore, we utilize a latent diffusion model to synthesize facial expressions with audio input. To further guarantee the accuracy and fidelity of the generated upper face movements, an Emotion Adapter is employed within the upper face latent space. The operational details and the efficacy of the conditional diffusion model alongside the Emotion Adapter's role are comprehensively expounded in Sections~\ref{Conditional Latent Diffusion Model} and~\ref{Conditional Latent Diffusion Model with Emotion Adapter} of our study. The complete pipeline of our approach is illustrated in Fig.~\ref{fig2}.

\subsection{Dual Facial Expression Representation}
\label{Dual Facial Expression Representation in Latents}

Since the movements of the eye region don't correspond one-to-one with audio like lip movements do, multiple eye movements can often appear plausible for the same vocal emotion, which cause problems when jugging the emotion. To solve these problems, we separate the face ($F$) into two parts: the upper face region ($f_1$), which contains eyebrows, eyelids, and eyeball movements, and the mouth region ($f_2$), which focuses on mouth movements (see Tab.~\ref{tab:Ablation results structure}). We propose the Variational AutoEncoder (VAE) for each part of the face seperately, denoted as $\mathcal{V} = \{v_1, v_2\}$, leveragin the transformer encoder $\mathcal{E} = \{e_1, e_2\}$ and the transformer decoder $\mathcal{D} = \{d_1, d_2\}$~\cite{petrovich2021action}. The expression VAE $\mathcal{V} = \{\mathcal{E}, \mathcal{D}\}$ is meticulously trained to reconstruct facial blenshapes sequences $B^{1:L} \in \mathbb{R}^{51 \times L}$, where $L$ represent the frame length of the blenshape sequence. We mask the facial blenshapes sequences into two parts $B^{1:L} = \{b_{1}^{1:L}, b_{2}^{1:L}\}$ according to the position on the face and take it as the corresponding VAE input. Our model integrates enhanced transformer modules in both $\mathcal{E}$ and $\mathcal{D}$, incorporating long skip connections~\cite{tevet2023human}, to facilitate robust feature encoding and decoding.

The encoder $\mathcal{E} = \{e_1, e_2\}$ is designed to ingest a learnable distribution token, serving as a pivotal element in the encoding phase. This token plays a critical role in extracting salient information from the input facial blenshapes sequences $B^{1:L}$ on a frame-by-frame basis. This approach captures the dynamic nature of facial expressions, enabling the encoder to produce a detailed representation of the input facial expression sequences. Following Kingma and Welling~\cite{kingma2013auto}, we reparameterize to obtain the latent variable $Z = \{z_1, z_2\}, z_1, z_2 \in \mathbb{R}^{n \times d}$, where we choose $n=1, d=256$ in our experiments, mirroring the dimensions outlined by Petrovich et al.~\cite{petrovich2021action}.
\begin{equation}
Z = \mathcal{E}(B^{1:L}) = \{ z_1 = e_1(b_{1}^{1:L}), z_2 = e_2(b_{2}^{1:L}) \}
\end{equation}
The latent variable $ Z \in \mathbb{R}^{n \times d} $ serves as the memory input of the cross-attention with an $L$ frame zero facial blenshape sequence token as queries in the decoder $\mathcal{D} = \{d_1, d_2\}$ process. This sophisticated setup allows $\mathcal{D}$ to generate a sequence of facial blendshapes $\hat{B}^{1:L}$, consisting of $L$ frames, showcasing the capability of the decoder to produce detailed and fluid facial expression sequences.

\begin{equation}
 \hat{B}^{1:L} = \mathcal{D}(Z) = \{d_1(z_1), d_2(z_2)\}
\end{equation}

In line with the findings by Petrovich et al.~\cite{petrovich2022temos}, the stability and complexity of generated facial expressions are significantly augmented through the strategic use of the latent space $Z$ and variable duration. To enhance the quality of the latent representations, a long skip-connection is implemented between the transformer-based encoder $\mathcal{E}$ and decoder $\mathcal{D}$ utilizing both the Mean Squared Error (MSE) loss and the Kullback-Leibler (KL) divergence loss. This innovation part ensures a more effective flow and integration of information within the model , facilitating the generation of diverse and nuanced facial expressions. We also explore the effectiveness of the latent's dimensions on facial expression sequences representation in Table 4. Hence, our VAE models present a stronger face reconstruction ability and richer diversity (see Tab.~\ref{tab:Ablation results else}).

\subsection{Conditional Latent Diffusion Model}
\label{Conditional Latent Diffusion Model}

Diffusion probabilistic models, as introduced by Sohl-Dickstein et al.~\cite{sohl2015deep}, present a methodology for progressively denoising a signal from a Gaussian distribution towards the facial expression distribution $p(b)$. This process involves learning to predict the noise across a $\mathcal{T}$-length Markov noising sequence, $\{b_{t}\}_{t=1}^{T}$. This paradigm has profoundly influenced various domains, notably in the advancement of image synthesis~\cite{dhariwal2021diffusion}, density estimation~\cite{kingma2021variational}, and motion generation models~\cite{tevet2023human}. Specifically for facial expression generation, models are trained using a transformer-based denoiser $\varepsilon_{\theta}(b_t, t)$, adept at converting random noise into sequences of facial expressions $\{\hat{b}_{1:N_t}\}_{t=1}^{T}$ through an iterative denoising process.

The direct application of diffusion models to facial expression sequences has been identified as inefficient and resource-intensive. Additionally, facial expression data, whether derived from markerless or marker-based systems, often contain high-frequency outliers that potentially degrade the model's learning efficacy. To address these challenges, our approach involves executing the diffusion process within a low-dimensional and representative latent space of facial expressions.

In this work, we adopt the denoising function $\varepsilon_{\theta}(B_t, t)$ from Chen et al.~\cite{chen2023executing}, pivoting from the traditional UNet architecture~\cite{ronneberger2015unet} towards a transformer-based denoising model, enhanced with long skip connections~\cite{bao2022all}. This model operates within the latent space $z \in \mathbb{R}^{n \times d}$, offering an architecture inherently more suitable for sequential data like facial expression sequences. We conceptualize the diffusion in latent space as a Markov denoising process, characterized by the transition probability:
\begin{equation}
q(Z_{t}|Z_{t-1}) = \mathcal{N}(\sqrt{\alpha_{t}} Z_{t-1}, (1-\alpha_{t})I),
\end{equation}
where $\alpha_{t} \in (0,1)$ acts as a hyper-parameter for the sampling process. The noise sequence $\{Z_{t}\}_{t=0}^{T}$ undergoes denoising at each step $t$, with $Z_{t-1} = \varepsilon_{\theta}(Z_{t}, t)$.

The process of generating facial expressions, denoted as $\mathcal{G}(a)$, is governed by the audio conditional distribution $p(Z|a)$. To refine this process, we employ a conditional denoiser, $\varepsilon_{\theta}(Z_{t}, t, a)$, which leverages a facial expression VAE model. The wav2vec audio encoder, represented as $\tau_\theta(a^{1:N})$ and producing embeddings in $\mathbb{R}^{N\times d}$~\cite{baevski2020wav2vec}, is implemented to efficiently map the audio embeddings. In integrating these embeddings into the transformer-based denoiser $\varepsilon_{\theta}$, we explore two primary methods: concatenation and cross-attention. Our empirical analysis, in line with Tevet et al.~\cite{tevet2023human}, reveals that concatenation is more effective, leading us to favor this method (see Tab. ~\ref{tab:Ablation results else}). The objective function for the conditional latent space is formalized as follows:
\begin{equation}
L_{\text{lat}} = \mathbb{E}_{\varepsilon, t, a} \left[ \left \|\varepsilon - \varepsilon_{\theta}(Z_{t}, t, \tau_\theta(a)) \right \|^2_2 \right]
\end{equation}
where $\varepsilon \sim \mathcal{N}(0, 1)$ and $Z_0 = \mathcal{E}(B_{1:L})$. During training, the encoder $\mathcal{E}$ remains fixed, encapsulating motion information into $Z_{0}$. Samples for the forward diffusion process are drawn from the distribution $p(Z_{0})$. In reverse diffusion, $\theta$ predicts $\hat{Z}_0$ across $T$ denoising steps, subsequently decoded by $\mathcal{D}$ into the final facial expression in a single forward pass.

\subsection{Emotion Adapter}
\label{Conditional Latent Diffusion Model with Emotion Adapter}

We introduce an encoder-only Emotion Adapter \(F_{apt}\) to evaluate the upper part facial movements generated in the latent space:
\begin{equation}
L_{\text{$F_{apt}$}} = -\frac{1}{N} \sum_{i=1}^{N} \left[ \log(\mathcal{D}(Z_{t})_{i}) \right],
\end{equation}

where \(Z_{t}\) represents the input latent sequence and \(\mathcal{D}\) decodes it into the corresponding facial movement sequence.

Our investigation reveals significant differences in the distribution patterns of facial expressions, particularly between the mouth and upper facial components. This finding contrasts with the approach of Meshtalk~\cite{richard2021meshtalk}, which categorizes facial expressions into two types: those correlated with audio (primarily movements of the mouth) and those uncorrelated (such as movements of the eyebrows and eyelids). We observe that, although audio features can accurately synchronize lip movements with corresponding audio waveforms, the emotional expressions in the upper face region follow a distinct pattern. This discrepancy is especially pronounced in spoken communication, where mouth movements are tightly coupled with speech phonetics, while the emotional expressions conveyed through the upper face region vary significantly based on the speaker's emotional state. 

In the animation industry, character emotions are often stereotypically conveyed through the eye region, particularly the eyebrows. To capture these distinct expressions, we introduce a novel dual latent space architecture in our facial expression Variational Autoencoder (VAE) model. This innovative approach, diverging from conventional single encoder-decoder VAE architectures, enables a more nuanced representation and diffusion of various facial expression sets.

The training objective for the upper-face denoiser combines both the latent diffusion loss \(L_{\text{lat}}\) and the Emotion Adapter loss \(L_{F_{apt}}\):

\begin{equation}
L_{\text{upper}} 
= \lambda_{\text{lat}} \, L_{\text{lat}} 
+ \lambda_{F_{apt}} \, L_{F_{apt}},
\end{equation}

where \(\lambda_{\text{lat}}\) and \(\lambda_{F_{apt}}\) are hyperparameters that balance the relative importance of each loss term. In our experiments, we set \(\lambda_{\text{lat}} = 1\) and \(\lambda_{F_{apt}} = 10\) (see Table~\ref{tab:Ablation results else}). Meanwhile, for the mouth region, only the latent diffusion loss is applied:

\begin{equation}
L_{\text{mouth}} 
= L_{\text{lat}} 
= \mathbb{E}_{\varepsilon, t, a} \bigl[\|\varepsilon - \varepsilon_{\theta}(Z_t, t, \tau_\theta(a))\|_2^2 \bigr].
\end{equation}

By incorporating the Emotion Adapter and carefully balancing these losses, our method effectively enforces emotional consistency in the upper-face region while preserving accurate lip synchronization. This design choice allows EmoDiffusion to generate high-fidelity, emotionally rich facial animations from audio inputs with minimal computational overhead.

\section{Dataset}

\begin{table}[!tbp]
\centering
\resizebox{0.85\textwidth}{!}{
\begin{tabular}{c|c|c|c|c|c}
\hline
Dataset & Language & Emotion Range & Data Format & Data Area & Duration \\ \hline
VOCASET~\cite{nocentini2024emovoca} & English & - & 5023 vertices & Lip & 30mins \\
BIWI~\cite{fanelli20103} & English & - & 23370 vertices & Lip & 1h \\
Multiface~\cite{wuu2022multiface} & English & - & 7306 vertices & Full face & 1.5h \\
3D-ETF(RAVDESS)~\cite{peng2023emotalk} & English & - & 52 Blendshapes & Full face & 2.5h\\
3D-ETF(HDTF)~\cite{peng2023emotalk}  & English & 8 & 52 Blendshapes & Full face & 3h \\
BEAT~\cite{liu2022beat} & English & 8 & 51 Blendshapes & Full face & 60h \\ \hline
\textbf{3D-BEF(ours)} & Chinese & 9 & 51 Blendshapes & Full face & 3h \\ \hline
\end{tabular}}
\caption{Dataset comparisons. We compare our 3D-BEF dataset with existing talking face datasets. Emotion Range refers to the emotion contained in the dataset audio. Data Area shows the collected data focus area.}
\label{tab:dataset_comparison}
\end{table}

\begin{figure}[!tbp]
\centering
\includegraphics[width=\textwidth, trim=30 280 30 0, clip]{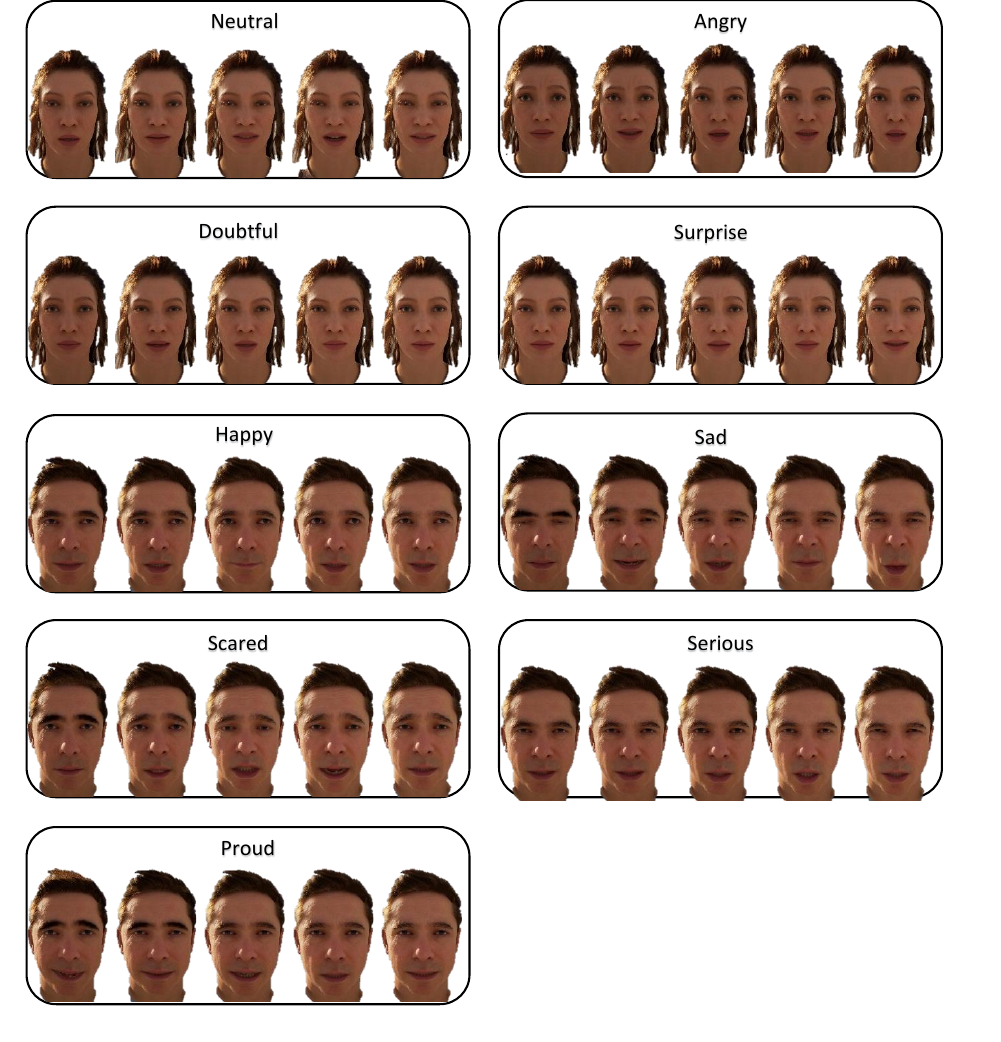}
\caption{3D-BEF dataset consists of nine different significant facial expression  with audio annotaions from different annotators. The figure showcases one example of each representative emotion of our dataset, including neutral, angry, doubtful, surprised, happy, sad, scared, serious, and proud.}
\label{fig data}
\end{figure}

To facilitate conditional facial expression synthesis, it is imperative to establish a comprehensive mapping between audio inputs and facial movements. Existing 3D datasets, such as BIWI~\cite{fanelli20103} and VOCASET~\cite{cudeiro2019capture}, while foundational, exhibit limitations for generative tasks. Most notably, they lack upper-face movements and a diverse range of emotional expressions. To address these limitations, we have compiled a medium-sized dataset encompassing a spectrum of emotional states, including neutral, angry, doubtful, surprised, happy, sad, scared, serious, and proud. The dataset content details shown in Fig.\ref{fig data}.

This dataset was captured using an iPhone 12 in conjunction with the LiveLinkFace application, enabling the acquisition of facial movements across 51 blendshapes dedicated to nuanced facial expressions. Our dataset comprises approximately 2000 facial expression sequences, each captured at a frame rate of 25 FPS and spanning roughly 4 seconds. To ensure a consistent evaluation framework, we employed identical splits for training, validation, and testing across all methods compared in this study, including CodeTalker~\cite{xing2023codetalker}, SelfTalk~\cite{peng2023selftalk}, classifier-free diffusion~\cite{tevet2023human}, Faceformer~\cite{fan2022faceformer}, and Facediffuser~\cite{stan2023facediffuser}. Specifically, we selected representative emotional scenes as test sets. Moreover, to verify the robustness and generalization of our model, we also used two open-source datasets, BEAT~\cite{liu2022beat} and 3D-ETF~\cite{peng2023emotalk}, and compared the results with state-of-the-art (SOTA) methods.

\section{Experiments}

In this section, we conduct a thorough comparative analysis to ascertain the efficacy and efficiency of our proposed models. We initiate with Subsection ~\ref{Implementation Details}, which discusses the detailed implementation aspects of our methodology. The outcomes of the evaluation and evaluation metrics employed are presented in Subsection ~\ref{Evaluation Results}, offering a comparison of our results with various state-of-the-art (SOTA) facial expression generation techniques, encompassing both quantitative evaluations and user studies. The final subsection, ~\ref{Ablation}, is dedicated to an ablation study that underscores the contribution of individual components to the overall performance of our model.

\subsection{Implementation Details}
\label{Implementation Details}

Our experimental framework was designed to evaluate the efficacy of our proposed facial latent diffusion model against contemporary state-of-the-art methodologies, including Code-Talker~\cite{xing2023codetalker}, SelfTalk~\cite{peng2023selftalk}, Faceformer~\cite{fan2022faceformer}, classifier-free diffusion~\cite{tevet2023human}, and Facediffuser~\cite{stan2023facediffuser}. To maintain the integrity of our comparative analysis, we utilized the same implementations of all the baseline models for both training and testing phases on our dataset, thereby ensuring a fair and consistent benchmarking process. For the BEAT~\cite{liu2022beat} and 3D-ETF~\cite{peng2023emotalk} datasets, we followed the official guides provided for these baseline models during both the training and testing phases to ensure consistency and fairness in our evaluations.

In our comparative analysis, the transformer encoders $\mathcal{E}$ and decoders $\mathcal{D}$, integral to the reconstruction VAE model $\mathcal{V}$, are characterized by a configuration of 9 layers and 4 heads. Skip connections are inherently integrated within this architecture. The transformer-based denoiser, denoted as $\varepsilon_{\theta}(Z_t, t)$, adheres to a structurally analogous framework. Concatenation is employed for the audio embedding $\tau_{\theta}(a) \in \mathbb{R}^{256}$ and the latent variable $Z \in \mathbb{R}^{256}$, facilitating the processes inherent in diffusion learning and inference. The facial action Emotion Adapter $d$ is built upon 4 transformer encoder layers with 4 heads, similarly incorporating skip connections as a default configuration. The conditional audio input leverages a pre-trained $Wav2Vec2-large-xlsr-53$ model, utilized here in a frozen state as the audio encoder. Our ablation studies, elucidated in Section ~\ref{Ablation}, scrutinize variables including but not limited to the Emotion Adapter weight, latent space dimensionality, and architectural depth. Model training employs the AdamW optimizer, maintaining a constant learning rate of $10^{-4}$. Mini-batch sizes are standardized at 32 across the VAE and diffusion model training phases. The training duration extends over 1,000 epochs for the VAE and is doubled to 2,000 epochs for the diffusion model training. Training protocols stipulate 1,000 diffusion steps, which are reduced to 50 during the inference phase. The variance parameters $\beta_{t}$ are adjusted to scale linearly from $8.5 \times 10^{-4}$ to $0.012$. Concerning computational resources and runtime, the training of VAEs $\mathcal{V}$ is executed over a span of 24 hours, whereas the denoiser $\varepsilon_{\theta}$ demands 48 hours of runtime, both scenarios utilizing a single NVIDIA A100 GPU. Similarly, the evaluation phase is conducted on an identical computational setup.

    \subsection{Evaluation Results}
\label{Evaluation Results}

\subsubsection{Evaluation Metrics}
Evaluation Metrics are categorized into three distinct aspects to assess the performance of our facial animation system. 

    
    

\begin{enumerate}
    \item \textbf{Facial Blendshape Error (FBE):} The synchronization of facial expressions is quantitatively evaluated through the Facial Blendshape Error (FBE). This metric, inspired by the methodologies in Codetalker~\cite{xing2023codetalker} and Selftalk~\cite{peng2023selftalk}, is defined as the mean L2 norm error between the predicted and actual facial blendshapes across a test set. FBE specifically assesses the accuracy of facial expression synchronization, with a particular focus on the lip region to ensure the precision of speech-related movements.
    
    \item \textbf{Emotional Blendshape Error (EBE):} To gauge the fidelity of emotional expressions mirrored in the audio sequence, we introduce the Emotional Blendshape Error (EBE). Drawing inspiration from the concept of emotional vertex error in Emotalk~\cite{peng2023emotalk}, EBE is calculated as the maximum L2 norm error among blendshape coordinates, emphasizing eyebrow region displacement. This metric is designed to capture both the intensity and accuracy of emotional expressions, underlining the significance of eyebrow movements in the effective conveyance of emotions.
    
    \item \textbf{Facial Dynamics Deviation (FDD):} Evaluating the deviation in upper-face dynamics is paramount for assessing the naturalness of facial animations. Following the approach in Selftalk~\cite{peng2023selftalk}, FDD measures the dynamic variations in facial expressions against the ground truth throughout a motion sequence. This metric accentuates the diversity in facial expressions associated with speech, which correlates loosely with the verbal content and style, reflecting considerable individual variability.
\end{enumerate}

These metrics collectively offer a holistic evaluation framework, enabling an in-depth analysis of both the accuracy and expressiveness of generated facial animations. The incorporation of both blendshape synchronization and emotional expressiveness ensures a balanced assessment of the system's capability to produce realistic and engaging animations.


\begin{table}[!tbp]
\centering
\caption{Result of Quantitaive evaluation. We compare Our methods with five state-of-art methods on our dataset. The average error is used for the evaluation}
\label{tab:Quantitative Evaluation}
\resizebox{0.85\textwidth}{!}{%
\begin{tabular}{lcccc}
\hline
Dataset & Method & FBE $^{\times10^{-2}}$ $\downarrow$ & EBE $^{\times10^{-2}}$ $\downarrow$ & FDD $^{\times10^{-4}}$ $\downarrow$ \\ \hline
\multirow{5}{*}{RAVDESS}& VOCA~\cite{nocentini2024emovoca} & 4.538 & 4.188 & -  \\    
& MeshTalk~\cite{richard2021meshtalk} & 3.414  & 3.386 & -    \\ 
& FaceFormer~\cite{fan2022faceformer}  & 3.560 & 3.757 & -    \\ 
& Emotalk~\cite{peng2023emotalk}  & 2.597 & \textbf{2.493} & -    \\ \cline{2-5}
& \textbf{Ours}   & \textbf{2.571} & 2.613 & - \\ \hline
\multirow{5}{*}{HDTF }& VOCA ~\cite{nocentini2024emovoca} & 3.735 & 3.286 & -  \\    
& MeshTalk~\cite{richard2021meshtalk} & 3.419 & 3.124 & -    \\ 
& FaceFormer~\cite{fan2022faceformer}  & 3.232 & 3.142 & -    \\ 
& Emotalk~\cite{peng2023emotalk}  & 2.568 & 2.364 & -    \\ \cline{2-5}
& \textbf{Ours}   & \textbf{2.141} & \textbf{2.215} & - \\ \hline
\multirow{3}{*}{BEAT}& FaceDiffuser ~\cite{stan2023facediffuser} & 5.152 & 13.58 & 14.71 \\
& FaceDiffuser~\cite{stan2023facediffuser} (w/o diffusion)   & 4.170 & 10.77 & 14.82 \\ \cline{2-5} 
& \textbf{Ours}   & \textbf{3.128}  & \textbf{8.03}  & \textbf{8.97}  \\ \hline
\multirow{5}{*}{3D-BEF} & CodeTalker ~\cite{xing2023codetalker} & 2.61 & 4.21 & 12.53 \\ 
& SelfTalk~\cite{peng2023selftalk} & \textbf{2.53} & 3.83 & 11.02 \\ 
& FacialDiffusion~\cite{tevet2023human} & 3.09 & 5.30 & 25.46 \\
& FaceFormer~\cite{fan2022faceformer}  & 3.69 & 7.35 & 9.78 \\
& FaceDiffuser~\cite{stan2023facediffuser} & 3.57 & 4.44 & 8.30 \\  \cline{2-5}
& \textbf{Ours} & 2.81 & \textbf{3.22} & \textbf{7.99} \\  \hline
\end{tabular}%
}
\end{table}

\subsubsection{Quantitative Evaluation}
Tab.~\ref{tab:Quantitative Evaluation} presents a comparative analysis of our method against a variety of state-of-the-art baselines, reporting three key metrics: \textit{Emotional Blendshape Error (EBE)}, \textit{Facial Dynamic Detail (FDD)}, and \textit{Facial Blendshape Error (FBE)}. Our results reveal that, while our approach does not obtain the lowest FBE, it delivers superior performance in terms of EBE and FDD across our own dataset.

The underlying reasons for this strong performance on EBE and FDD merit closer examination. First, although diffusion-based generative models can introduce variability at each sampling step, potentially undermining blendshape precision, our design incorporates an additional \textit{Emotion Adapter} to counteract such variability. This module specifically refines emotional blendshape sequences by improving emotional fidelity and enhancing the dynamic details of facial animations. The outcome is a system that excels in capturing the finer nuances of facial expressions (e.g., subtle muscle movements around the eyes and eyebrows) and in depicting dynamic facial transitions, resulting in more expressive and lifelike animations.

Furthermore, our method consistently outperforms baseline approaches on both the BEAT dataset and the 3D-ETF dataset. Specifically, on the BEAT dataset, it achieves the highest overall scores among all metrics reported in~\cite{stan2023facediffuser}. Turning to the emotion-centric RAVDESS dataset~\cite{livingstone2018ryerson}, which comprises just two sentences with varying degrees of emotional intensity, our audio-only conditioned solution outperforms others in FBE but is marginally surpassed by EmoTalk~\cite{peng2023emotalk}, likely because EmoTalk explicitly conditions on emotion-level labels. In contrast, for the HDTF dataset~\cite{zhou2021vaw}, which focuses on realistic talking scenarios, our approach ranks first among state-of-the-art methods across all reported metrics.

Taken together, these results underscore the advantages of blending diffusion-based generative modeling with targeted discriminative refinements. By using the Emotion Adapter to improve emotional accuracy and enhance dynamic details, we establish a new performance benchmark in emotional and dynamic facial animation.

\begin{figure}[!tbp]
\centering
\includegraphics[width=\textwidth, trim=0 100 50 0,clip]{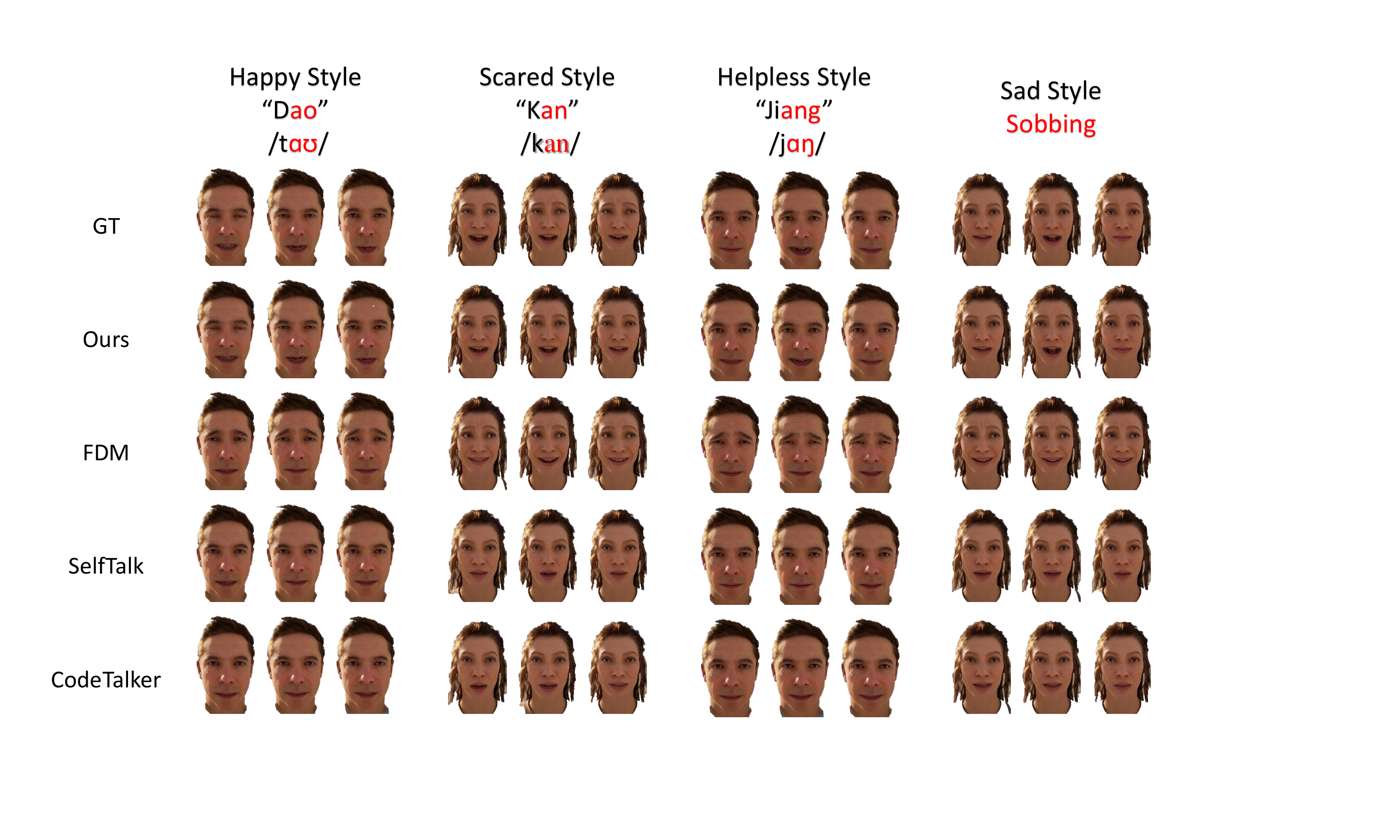}
\caption{We conducted a visual comparison of facial movements using different methods trained on the 3D-BEF dataset. Our analysis focused on various syllables, revealing that our method achieved greater precision in mouth movements and upper face expressions across different emotional styles. Notably, our method exhibited enhanced movement in the eyebrows and eyelids, effectively capturing the essence of a happy emotion, exemplified by the syllable \textipa{"/tA\textupsilon/"}. Furthermore, for syllables necessitating mouth curving, such as \textipa{"/jA\ng{}/"}, our approach demonstrated superior accuracy in depicting the curving mouth movements. Our methods also express emotions well with modal particles like sobbing.}

\label{fig3}
\end{figure}

\subsubsection{Perceptual evaluation}
Audio and facial movements are inherently multifaceted phenomena that transcend simple quantitative metrics, necessitating a nuanced approach to evaluation that incorporates human perceptual analysis. To this end, we undertook a qualitative assessment of our model, delineating its performance from three pivotal perspectives:

\begin{enumerate}
    \item \textbf{Lip Synchronization:} While previous works have made significant strides in lip synchronization, a critical analysis reveals a gap in capturing the nuanced alignments with human speech emotions. In contrast, our model, when benchmarked against Codetalker~\cite{xing2023codetalker} and Selftalk~\cite{peng2023selftalk} under identical audio conditions, demonstrates superior lip movement articulation and a more accurate reflection of speech-driven emotional nuances. As illustrated in Fig.~\ref{fig3}, our approach excels in rendering complex mouth movements associated with speech. For instance, for syllables requiring a wide mouth opening, such as \textipa{"/tA\textupsilon/"}, our model enhances the perception of the spoken content through improved mouth aperture. Similarly, for syllables necessitating mouth curving, such as \textipa{"/jA\ng{}/"}, our model achieves superior accuracy in depicting these intricate movements compared to previous state-of-the-art methods. 

    \begin{figure}[!tbp]
    \centering
    \includegraphics[width=\textwidth, trim=0 50 0 0, clip]{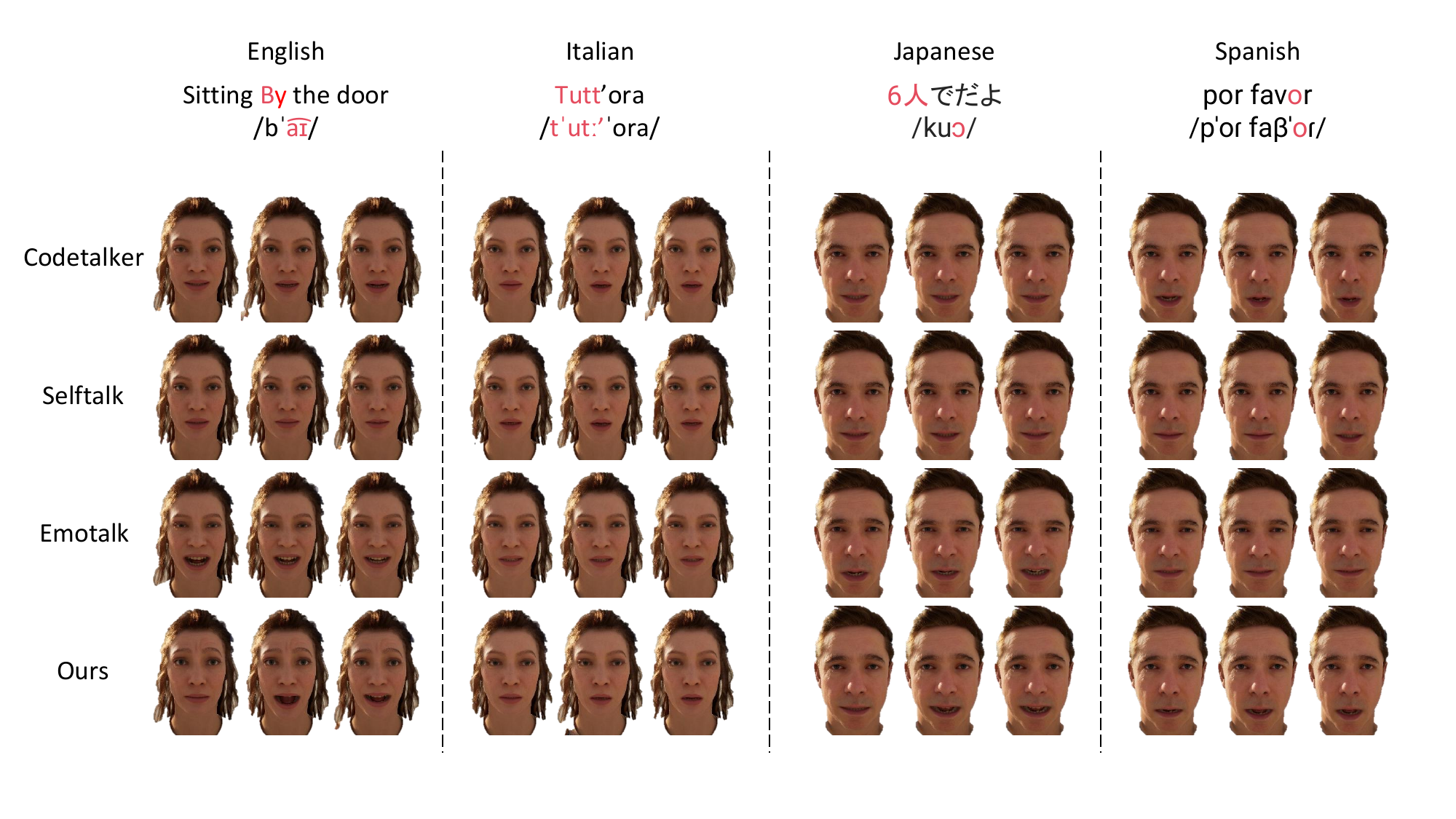}
    \caption{Incorporating audio input from various languages, EmoDiffusion produces natural and dynamic 3D facial animation sequences, maintaining clarity even in fast-talking scenarios, and accurately synchronizing both lip movements and emotional cadences of speech.}
    \label{fig5}
    \end{figure}
    
    \item \textbf{Langauge Generalizability} As illustrated in Fig.~\ref{fig5}, our method outperforms the state-of-the-art Emotalk~\cite{peng2023emotalk} in various languages. Particularly in fast-talking scenarios, such as the left Spanish sequences, our model maintains clear movements, while Emotalk appears static when pronouncing the "o" in \textipa{"/porfa$\beta$or/"}. These improvements not only underscore our model's advanced lip-sync capabilities but also highlight its proficiency in synchronizing with the underlying emotional cadences of speech, thereby enhancing the realism and relatability of the generated facial animations.

    \item \textbf{Emotional Expression:} The limitations of previous methods were primarily due to their optimization for static rather than dynamic emotional expressions, resulting in a notable deficiency in accurately reflecting the range of emotional tones. In contrast, our approach focuses on generalizing the dynamic nature of emotion-reflective actions within audio sequences. As illustrated in Fig.~\ref{fig3}, when expressing happiness, the raised eyebrows and upward corners of the mouth distinctly convey joy. This enhancement allows for a vivid portrayal of emotions communicated through speech, lending a more realistic and natural appearance to the facial expressions. The effectiveness of our model is further demonstrated in the accompanying video supplement, which provides a detailed comparison and showcases the model's ability to render emotional expressions with remarkable fidelity.
\end{enumerate}

\begin{figure}[!tbp]
\centering
\includegraphics[width=0.85\textwidth,trim=100 150 100 100,clip]{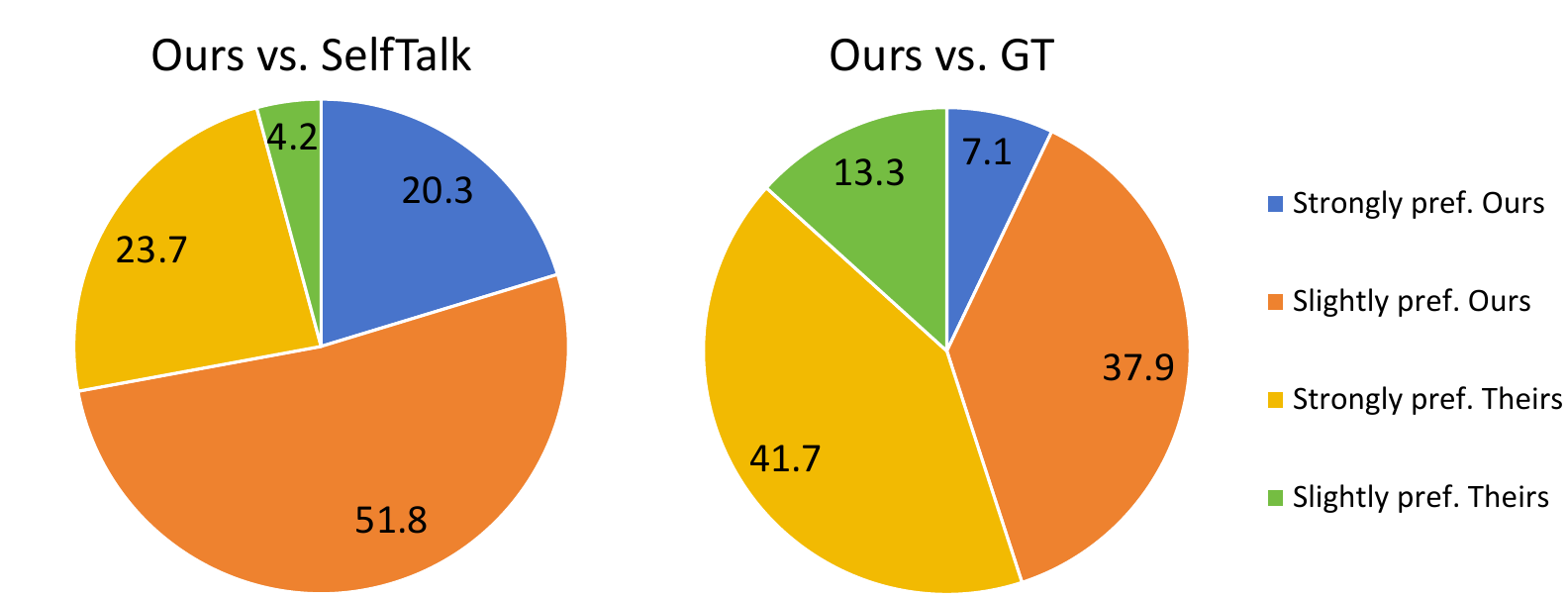}
\caption{Perceptual evaluation on Ours vs. Ground Truth or Ours vs. SelfTalk~\cite{peng2023selftalk}. The evaluation focused on the mesh visualization results of facial expressions generated by each method. Remarkably, participants in the study found it challenging to distinguish between the facial expressions generated by our model and the ground truth.}
\label{fig4}
\end{figure}

We conduct a perceptual evaluation with approximately 50 participants to assess the effectiveness of our method compared to the leading baseline, Selftalk~\cite{peng2023selftalk}, and the ground truth. We employ A/B testing for each comparison, \textit{i.e.}, our method versus the competitor, focusing on realistic facial animation and lip synchronization. As depicted in Fig.~\ref{fig4}, our method significantly outperforms Selftalk, with over 70\% of evaluators preferring our mesh visualizations. Moreover, our approach performs competitively close to the ground truth in visual assessments, with around 45\% of evaluators favoring our method over the ground truth. These results demonstrate the efficacy of our approach. For detailed comparisons, we encourage readers to refer to the supplementary videos provided.

The foregoing analysis elucidates our model's distinctive advantages in achieving lifelike and emotionally resonant facial animations, setting a new benchmark in the domain of audio-driven facial animation technology.

\subsection{Ablation}
\label{Ablation}

In this section, we delve into an ablation study designed to dissect the contributions of distinct components within our proposed facial expression latent diffusion framework, particularly in relation to the fidelity of the synthesized 3D talking faces. Our investigation is multifaceted, beginning with an examination of the framework's performance in the absence of the Emotion Adapter, a pivotal element tasked with refining the generated facial expressions to ensure realism and coherence with the input audio cues. Following this, we explore the influence of varying the structure of the latent space, a critical factor that potentially affects the model's capacity to capture and represent complex emotional nuances and facial dynamics. 

\begin{table}[!tbp]
\centering
\footnotesize
\caption{Result of the impact of different structure of latent space (\# L) and with(w/) or without(w/o) Emotion Adapter ($F_{apt}$). We use metrics in Tab.\ref{tab:Quantitative Evaluation} and provides real reference, the 1 indicate the first dimension of the latent variable.}
\label{tab:Ablation results structure}
\resizebox{0.85\textwidth}{!}{%
\begin{tabular}{lccccc}
\hline
Model & $F_{apt}$ & \# L & FBE $^{\times 10^{-2}}$ $\downarrow$ & EBE $^{\times 10^{-2}}$ $\downarrow$ & FDD $^{\times 10^{-4}}$ $\downarrow$ \\ \hline
Ours-1 & \checkmark & one  & 3.45 & 4.69 & 10.63  \\
Ours-1 & $\times$ & one  & 3.31 & 3.83 & 9.23 \\
Ours-1 & \checkmark & two  & 3.03 & 3.42 & 10.43 \\
\textbf{(Ours-1)} & $\times$ & one  & \textbf{2.81}& \textbf{3.22} &\textbf{7.99} \\  \hline
\end{tabular}%
}
\end{table}

\subsubsection{Impact of the Emotion Adapter}
We conduct a comprehensive investigation into the role of the Emotion Adapter by examining its impact on aligning generated facial animations with ground-truth motion curves. As shown in Table~\ref{tab:Ablation results else}, removing the Emotion Adapter results in a marked performance degradation across all evaluated metrics. The most pronounced effect is the substantial increase in the Facial Dynamic Deviation (FDD) from $7.99$ to $10.43$. This measure quantifies how closely the generated motions follow the temporal and amplitude-based variations observed in real human facial expressions. Such a large jump in FDD underscores the central importance of the Emotion Adapter in capturing dynamic aspects of facial expressions, particularly in the upper facial region, where subtle nuances (e.g., eyebrow raises, forehead wrinkles) convey crucial emotional content.

Further analysis reveals that the absence of the Emotion Adapter leads to a notable reduction in expression amplitude, especially within the upper facial region. This reduced amplitude directly hinders the model's ability to represent realistic human expressions. As a consequence, the overall fidelity of the generated animations decreases, making them appear more rigid or subdued. Crucially, our experiments show that losing the Emotion Adapter introduces an error more than ten times greater than that of our baseline configuration, highlighting its indispensable role in faithfully reconstructing the ground-truth motion curve.

In an effort to mitigate the severe performance drop caused by removing the Emotion Adapter, we explored different ratios between $\lambda_{\text{lat}}$ (weight for the latent loss) and $\lambda_{\text{$F_{apt}$}}$ (weight for the Emotion Adapter loss). The results, detailed in Table~\ref{tab:Ablation results else}, indicate that a ratio of $0.1$ yields notably improved performance over other tested ratios, reaffirming the vital function of the Emotion Adapter in preserving emotional detail and overall expressiveness in the generated animations.

\subsubsection{Impact of the Structure of the Latent Space} 
At the core of our framework lies the integration of a dual Variational Autoencoder (VAE) architecture, which operates in conjunction with a latent training diagram. The structure of the latent space plays a pivotal role in the model's performance. Through extensive experimentation, we observed a notable performance drop when the latent space was simplified into a single entity. This simplification caused significant degradation across key metrics. Specifically, the Facial Blendshape Error (FBE) increased from $2.81$ to $3.31$, and the Emotional Blendshape Error (EBE) rose from $3.22$ to $3.83$, as shown in Table~\ref{tab:Ablation results structure}. These results highlight the challenges inherent in reconciling the domain gaps between audio input and facial feature output, which are intensified when the latent space is reduced to a single modality.

This degradation suggests that data-driven reconstruction and latent diffusion models face considerable difficulty in learning accurate facial expressions when they are unable to disentangle the distinct features of audio and facial modalities. The ambiguity introduced by this modality fusion hampers the model's ability to capture precise emotional cues, which are essential for creating realistic facial animations. Our findings demonstrate that maintaining a well-structured, multi-dimensional latent space is crucial for achieving high-fidelity results, particularly in the accurate replication of complex facial expressions that are closely tied to the emotional content of the input speech.

\subsubsection{Effectiveness of the Denoiser in Conditional Latent-based Diffusion Models} 

We further investigate the design of our conditional latent diffusion framework by assessing multiple aspects of the denoiser configuration. Specifically, we vary (1) the shape of the latent representation, (2) the method used to fuse audio conditions into the diffusion process, and (3) the number of layers in the denoiser.

First, we examine latent vectors of different shapes $z \in \mathbb{R}^{i \times 256}$, as summarized in the first part of Table~\ref{tab:Ablation results else}. Our results suggest that the most compact form, $z \in \mathbb{R}^{1 \times 256}$, performs best on most evaluation metrics. This indicates that a smaller latent size can preserve the necessary expressiveness for facial animation while simultaneously reducing noise and complexity in the diffusion process. Consequently, a compact latent space appears more conducive to efficient training and inference, without sacrificing quality or diversity of the generated expressions.

Next, we analyze two alternative approaches for incorporating audio conditions into the diffusion process: cross-attention (\textit{cross-att}) and concatenation (\textit{concate}). In alignment with prior studies, such as MDM~\cite{tevet2023human} and MLD~\cite{chen2023executing}, our experiments confirm that concatenating audio embeddings generally yields superior outcomes compared to using cross-attention. We hypothesize that concatenation offers a more direct integration of the audio features, thus better preserving the temporal and amplitude information necessary for accurate lip synchronization and emotional cues.

Finally, in the third part of Table~\ref{tab:Ablation results else}, we explore how varying the number of layers in the denoiser $\epsilon_{\theta}$ influences performance. While the differences among configurations are relatively minor, a 9-layer denoiser appears to achieve marginally better results than either fewer or more layers. This suggests that an optimal balance exists between depth and overfitting risk, where too few layers may limit the capacity to model intricate temporal patterns, and too many layers may introduce unnecessary complexity and slow down training without appreciably enhancing quality.

Together, these findings attest to the robust capabilities of our conditional latent diffusion framework in synthesizing high-quality facial expressions. A compact latent dimension ($z \in \mathbb{R}^{1 \times 256}$), a concatenation-based fusion of audio embeddings, and an appropriately sized denoiser (9 layers) collectively bolster the effectiveness of our approach. By systematically refining the key components of the latent diffusion process, we achieve a more efficient model that excels in generating faithful, dynamic, and emotionally rich facial animations under diverse audio conditions.

\begin{table}[!tbp]
\centering
\caption{Result of the evaluation on our 3D-BEF dataset: we use metrics same as Tab. \ref{tab:Quantitative Evaluation}, we compare the different shape of latent $Z$, cross-attention or concatenation with condition $\tau_\theta$, different number of transformer layers in $\epsilon_{\theta}$,  and different weight ratios between $\lambda_{lat}$ and $\lambda_{F_{apt}}$}
\label{tab:Ablation results else}
\resizebox{0.85\textwidth}{!}{%
\begin{tabular}{lccc}
\hline
Model & FBE $^{\times 10^{-2}}$ $\downarrow$ & EBE $^{\times 10^{-2}}$ $\downarrow$ & FDD $^{\times 10^{-4}}$ $\downarrow$ \\ \hline
Ours-1 ($Z$, $\mathbb{R}^{1 \times 256}$) & \textbf{2.81}& 3.22 &\textbf{7.99} \\
Ours-3 ($Z$, $\mathbb{R}^{3 \times 256}$) & 3.60 & 4.84 & 9.74 \\
Ours-5 ($Z$, $\mathbb{R}^{5 \times 256}$) & 3.45 & \textbf{3.14} & 9.90 \\ \hline
\hline
Ours-1 ($\tau_\theta$, cross-att) & 3.48 & 4.91 & 9.30 \\
Ours-1 ($\tau_\theta$, concat) & \textbf{2.81}& \textbf{3.22} &\textbf{7.99} \\ \hline
\hline
Ours-1 ($\epsilon_{\theta}$, 7 layers) & 3.23 & 4.76 & 9.32 \\
Ours-1 ($\epsilon_{\theta}$, 9 layers) & \textbf{2.81}& 3.22 &\textbf{7.99} \\
Ours-1 ($\epsilon_{\theta}$, 11 layers) & 3.13 & \textbf{2.95} & 9.36 \\ \hline
\hline
Ours-1 ($\lambda_{lat}$: $\lambda_{F_{apt}}$, 10) & 3.02 & 5.51 &\textbf{6.87} \\
Ours-1 ($\lambda_{lat}$: $\lambda_{F_{apt}}$, 1) & 2.95 & 4.75 & 7.35 \\
Ours-1 ($\lambda_{lat}$: $\lambda_{F_{apt}}$, 0.1) & \textbf{2.81} & \textbf{3.22} & 7.99 \\ \hline
\end{tabular}%
}
\end{table}

\subsection{Limitation and Discussion}

Our method is capable of generating results of any length; however, the quality may decline when exceeding the maximum length observed during training. The use of latent space in the diffusion process introduces complexities in fusion operations that are more challenging compared to working in explicit motion space. Our dataset has been carefully designed to meet the standards of the animation industry, with input from professional animators to ensure its relevance and quality. In the animation industry, emotions are often conveyed in a stylized manner, primarily through the eye region, particularly the eyebrows, while the rest of the face plays a less significant role in reflecting emotions. Notably, our dataset is the first to focus specifically on upper facial features, especially eyebrow movements. Our approach emphasizes capturing the subtle movements of the actor's eyebrows and eyelids while intentionally minimizing exaggerated mouth movements, aligning with the unique style of animation production. While our method effectively addresses these challenges, as shown in this section, there is still room for improvement in making emotional expressions more nuanced. Future work may need to incorporate additional factors such as facial muscle movements to further enhance the realism and appropriateness of emotional expressions.

\section{Conclusions}

In this paper, we introduce EmoDiffusion, a novel framework that combines Variational Autoencoders with Emotion Adapter-enhanced diffusion models to create stereotypically, emotionally expressive 3D facial animations synchronized with speech. We specifically crafted the 3D Blendshape Emotional Talking Face Dataset (3D-BEF) for this project, advancing facial animation research by offering a diverse range of emotional expressions in an animation style under the guidance of professional animators. Our approach significantly reduces the resources required to produce high-quality animations, democratizing access for a broader range of creators and enhancing interactive multimedia applications. As digital interactions become increasingly sophisticated, EmoDiffusion is poised to shape the future of multimedia, making virtual characters more engaging and accessible.





\bibliographystyle{elsarticle-num} 
\bibliography{main}

\end{document}